\documentclass[sigconf]{acmart}
\DeclareUnicodeCharacter{00D7}{\ensuremath{\times}}
\usepackage{cleveref}
\usepackage{booktabs}  
\usepackage{makecell}
\usepackage{graphicx}
\usepackage{enumitem}

\usepackage{epsfig}
\usepackage{epstopdf} 
\usepackage{array}
\crefname{figure}{Fig.}{Figs.}
\Crefname{figure}{Figure}{Figures}

\AtBeginDocument{%
  }

\setcopyright{acmlicensed}
\copyrightyear{2025}
\acmYear{2025}
\setcopyright{cc}
\setcctype{by}
\acmConference[MM '25]{Proceedings of the 33rd ACM International Conference on Multimedia}{October 27--31, 2025}{Dublin, Ireland}
\acmBooktitle{Proceedings of the 33rd ACM International Conference on Multimedia (MM '25), October 27--31, 2025, Dublin, Ireland}\acmDOI{10.1145/3746027.3758208}
\acmISBN{979-8-4007-2035-2/2025/10}

\begin{document}

\title{VER-Bench: Evaluating MLLMs on Reasoning with Fine-Grained Visual Evidence}


\author{Chenhui Qiang}
\authornote{Both authors contributed equally to this paper.}
\affiliation{%
  \institution{University of Chinese Academy of Sciences}
  \city{Beijing}
  \country{China}
}
\email{qiangchenhui23@mails.ucas.ac.cn}

\author{Zhaoyang Wei}
\authornotemark[1]
\affiliation{%
  \institution{University of Chinese Academy of Sciences}
  \city{Beijing}
  \country{China}
}
\email{weizhaoyang23@mails.ucas.ac.cn}

\author{Xumeng Han}
\authornotemark[1]
\affiliation{%
  \institution{University of Chinese Academy of Sciences}
  \city{Beijing}
  \country{China}
}
\email{hanxumeng19@mails.ucas.ac.cn}

\author{Zipeng Wang}
\affiliation{%
  \institution{University of Chinese Academy of Sciences}
  \city{Beijing}
  \country{China}
}
\email{wangzipeng22@mails.ucas.ac.cn}

\author{Siyao Li}
\affiliation{%
  \institution{University of Chinese Academy of Sciences}
  \city{Beijing}
  \country{China}
}
\email{lisiyao24@mails.ucas.ac.cn}

\author{Xiangyuan Lan}
\affiliation{%
  \institution{Peng Cheng Laboratory}
  \city{Guangdong, shenzhen}
  \country{China}
}
\email{xiangyuanlan@life.hkbu.edu.hk}

\author{Jianbin Jiao}
\affiliation{%
  \institution{University of Chinese Academy of Sciences}
  \city{Beijing}
  \country{China}
}
\email{jiaojb@ucas.ac.cn.}

\author{Zhenjun Han}
\affiliation{%
  \institution{University of Chinese Academy of Sciences}
  \city{Beijing}
  \country{China}
}
\email{hanzhj@ucas.ac.cn}
\authornote{Corresponding author.}

\renewcommand{\shortauthors}{Chenhui Qiang et al.}

\begin{abstract}
   With the rapid development of MLLMs, evaluating their visual capabilities has become increasingly crucial. Current benchmarks primarily fall into two main types: basic perception benchmarks, which focus on local details but lack deep reasoning (\emph{e.g.}, "what is in the image?"), and mainstream reasoning benchmarks, which concentrate on prominent image elements but may fail to assess subtle clues requiring intricate analysis. However, profound visual understanding and complex reasoning depend more on interpreting subtle, inconspicuous local details than on perceiving salient, macro-level objects. These details, though occupying minimal image area, often contain richer, more critical information for robust analysis. To bridge this gap, we introduce the VER-Bench, a novel framework to evaluate MLLMs' ability to: 1) identify fine-grained visual clues, often occupying, on average, just 0.25\% of the image area; 2) integrate these clues with world knowledge for complex reasoning. Comprising 374 carefully designed questions across Geospatial, Temporal, Situational, Intent, System State, and Symbolic reasoning, each question in VER-Bench is accompanied by structured evidence: visual clues and question-related reasoning derived from them. VER-Bench reveals current models' limitations in extracting subtle visual evidence and constructing evidence-based reasoning chains, highlighting the need to enhance models' capabilities in fine-grained visual evidence extraction, integration, and reasoning for genuine visual understanding and human-like analysis. The dataset is available at https://github.com/verbta/ACMMM-25-Materials. 
\end{abstract}

\begin{CCSXML}
<ccs2012>
   <concept>
       <concept_id>10002944.10011123.10011130</concept_id>
       <concept_desc>General and reference~Evaluation</concept_desc>
       <concept_significance>500</concept_significance>
       </concept>
   <concept>
       <concept_id>10010147.10010178.10010224</concept_id>
       <concept_desc>Computing methodologies~Computer vision</concept_desc>
       <concept_significance>300</concept_significance>
       </concept>
 </ccs2012>
\end{CCSXML}

\ccsdesc[500]{General and reference~Evaluation}
\ccsdesc[300]{Computing methodologies~Computer vision}

\keywords{Visual Evidence Reasoning, Multimodal Large Language Models, Fine-grained Perception, Visual Reasoning}

\maketitle

\section{Introduction}
\begin{figure*}[h]
  \includegraphics[width=\textwidth]{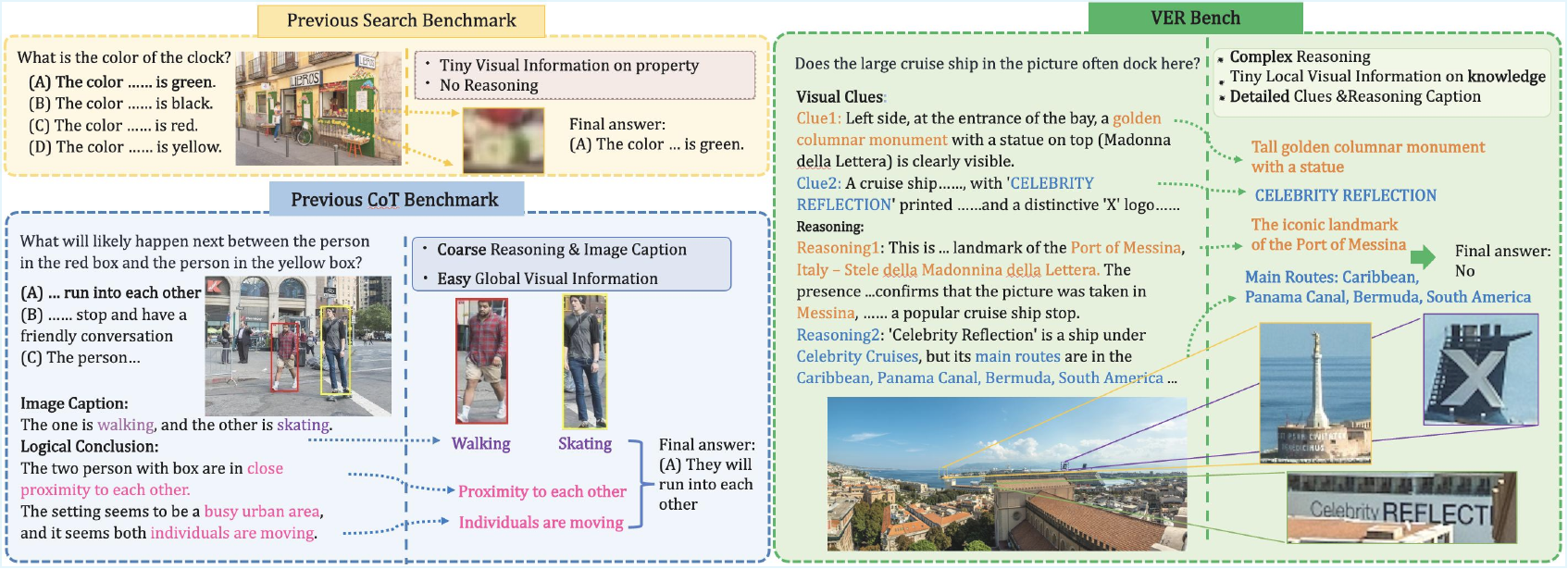}
  \caption{The difference between VER-Bench and existing visual evaluation benchmarks. {\mdseries The top-left benchmark detects fine visual details but lacks deep reasoning. The bottom-left benchmark incorporates reasoning yet relies on global image features with imprecise visual descriptions. The right side demonstrates our benchmark's objectives: (1) fine-grained local feature information, (2) detailed reasoning generation through visual clue.}}
  
  \label{fig:1}
\end{figure*}
The rapid evolution of multimodal large language models (MLLMs) has unlocked impressive capabilities in understanding and generating content that seamlessly blends vision and language~\cite{wu2024next, wang2024multimodal, li2024improving}. These models can now describe images with remarkable accuracy~\cite{bucciarelli2025, kim2025multi}, answer visual questions with nuanced responses~\cite{hu2024bliva, jian2024large}, and even generate images based on textual prompts~\cite{song2024moma, feng2025posellava}. However, a critical frontier remains largely unexplored: assessing whether these models genuinely reason by meticulously extracting and integrating specific visual evidence, or if their inferences sometimes rely more on learned textual priors or a more holistic understanding of obvious visual features. This issue is particularly pertinent when tackling complex inferences that require delving into subtle details, rather than merely leveraging high-level visual clues.

While several benchmarks now evaluate the visual capabilities of MLLMs, their focus often deviates from the specific demands of rigorous, locally evidence-based visual reasoning that VER-Bench aims to assess. As illustrated in the left sub-figure of Fig.\ref{fig:1}, certain benchmarks designed for basic visual perception, such as VStar~\cite{wu2024v} and MME-RealWorld~\cite{fu2024mme}, primarily emphasize identifying "what is in the image," focusing on local, minute details. However, they typically lack tasks that require deeper inferential reasoning grounded in those details. On the other hand, benchmarks incorporating Chain-of-Thought (CoT) methodologies, such as MME-CoT~\cite{jiang2025mme} and Visual CoT~\cite{shao2024visual}, aim to model reasoning processes. The visual grounding for these reasoning steps often relies on salient or holistic aspects of the image, rather than explicitly identifying subtle, fine-grained visual evidence. 

% This motivates our focus on small image regions, which can offer richer and more definitive information than larger, visually dominant areas. For instance, in the right sub-figure of Fig.\ref{fig:1}, when inferring the location depicted in the image, the prominent foreground skyline and large clusters of buildings may offer limited diagnostic value compared to a distant statue in the bay, which, acting as a landmark, can provide precise geographical clues. Similarly, when identifying a cruise ship, high-level visual features such as the vessel's size or color scheme may not provide unambiguous identification. In contrast, a detailed examination of small regions—such as a visible hull number or a unique logo—can lead to definitive recognition.

This motivates our focus on small image regions, which often provide more decisive information than larger, salient areas. For example, in the right sub-figure of Fig.\ref{fig:1}, the distant statue in the bay serves as a distinctive landmark for precise localization, whereas the prominent skyline and building clusters offer limited diagnostic value. 
Similarly, while a cruise ship’s size or color provides ambiguous cues, fine details such as a hull number or unique logo enable definitive identification.

To bridge this evaluation gap, we introduce VER-Bench, the Visual Evidence-based Reasoning Benchmark. This benchmark is specifically designed to assess an MLLM's ability to first extract minute yet crucial visual clues from images, and then construct a logical reasoning chain directly grounded in this granular evidence to support its conclusions. VER-Bench encompasses six challenging reasoning domains, each crafted to mirror diverse real-world analytical tasks:

- Geospatial Reasoning: Inferring location based on subtle geographic or cultural indicators.

- Temporal Reasoning: Determining time periods or sequences from visual context.

- Situational Adaptation Reasoning: Assessing appropriate actions or understanding evolving scenarios.

- Intent Inference: Deducing the purpose or goals of agents within the scene.

- System State Understanding: Analyzing the operational status or condition of depicted systems.

- Symbolic Analysis: Interpreting abstract symbols, codes, or non-linguistic visual information.

Each instance in VER-Bench is meticulously annotated with detailed visual evidence pairs, linking specific, often small and localized visual clues directly to the reasoning steps they support. This design mirrors sophisticated human reasoning, demanding that models identify subtle visual details—such as faint architectural patterns or small text—and construct verifiable, evidence-based reasoning chains. Each question is structured around a reasoning chain that includes: \textbf{(1) Precise Visual Clues} (\emph{e.g.}, small textual signs, symbols, minute environmental details) and \textbf{(2) Step-by-Step Reasoning} that logically connects these clues to the conclusion. The primary contributions of this work are summarized as follows:

\begin{itemize}[leftmargin=*]
\item Conceptualization of Visual Evidence-Based Reasoning: We formalize visual reasoning grounded in fine-grained evidence, highlighting its importance for rigorous MLLM evaluation.
\item Introduction of VER-Bench: We introduce a novel benchmark with tailored metrics to assess both final answer accuracy and the fidelity of evidence/reasoning chains.
\item Building a Scalable Data Pipeline: We detail a systematic pipeline for generating high-quality, diverse datasets to support scalable benchmark expansion.
\end{itemize}

\section{Related Work}
\subsection{Multimodal Large Language Models}

The development of multimodal large language models (MLLMs)~\cite{liu2023llava,zhu2023minigpt,lin2023sphinx,wang2024qwen2} represents a significant advancement in AI, building upon Large Language Models (LLMs)~\cite{touvron2023llama,qvq-72b-preview} and vision models~\cite{Radford2021LearningTV}. While closed-source models like GPT-4o~\cite{openai2024gpt4o} demonstrate strong capabilities, the research community has actively developed open-source alternatives such as LLaVA~\cite{liu2023llava}, MiniGPT-4~\cite{zhu2023minigpt}, InternVL2~\cite{chen2024far}, and Qwen2-VL~\cite{wang2024qwen2}. These advancements are driven by improved multimodal data~\cite{chen2025sharegpt4v,liu2024improved,wang2023see,ye2023mplug}, novel alignment techniques~\cite{bai2023qwenvl,dong2024internvl2,liu2024sphinx,li2024monkey,wang2023cogvlm}, and methods like LoRA~\cite{hu2021lora}. Recent research has focused on enhancing reasoning capabilities, with models like o1~\cite{o1} and works on multi-step Chain-of-Thought (CoT) reasoning with self-reflection~\cite{qvq-72b-preview,du2025virgo}. Despite increased accessibility through commercial APIs such as GPT-4v~\cite{achiam2023gpt4} and Gemini-Pro-V~\cite{team2023gemini}, questions regarding real-world efficacy remain~\cite{mme_realworld,mme}. 
% Concurrently, Unified MLLMs (U-MLLMs)~\cite{qvq-72b-preview,fu2025vita,yu2025aligning,zhang2024beyond} have emerged, targeting end-to-end multimodal understanding and generation. These models often employ unified training objectives~\cite{wang2024mio,wu2024vila,team2024chameleon}, explore vision encoding strategies like VQVAE for generation and SigLIP for comprehension~\cite{chen2025janus}, or utilize diffusion-based training for enhanced image generation~\cite{xie2024show,zhou2024transfusion}.

\subsection{MLLM Benchmarks}
The rapid advancement of MLLMs has spurred the creation of comprehensive evaluation benchmarks~\cite{duan2024vlmevalkit, fu2024mme}. These benchmarks cover a wide range of capabilities, including real-world applications and visual perception (\emph{e.g.}, MME~\cite{fu2024mme}, MMBench~\cite{liu2023mmbench}, MMVet~\cite{mmvet}), integrated reasoning (\emph{e.g.}, MMStar~\cite{zhang2024mmstar}, MMMU~\cite{yue2024mmmu}), and video understanding (\emph{e.g.}, MMBench-Video~\cite{fang2025mmbench}). Specialized benchmarks also exist for tasks like chart interpretation (ChartQA~\cite{masry2022chartqa}), document analysis (DocVQA~\cite{mathew2021docvqa}), and mathematical reasoning~\cite{yan2024errorradar}. Evaluation methods primarily include Visual Question Answering (VQA), which allows open-ended responses but faces scoring objectivity issues~\cite{eval:bleu,eval:cider,eval:spice}, and Multiple Choice Questions (MCQ)~\cite{data:A-okvqa}, which are prevalent in leaderboards due to more reliable outcomes. However, existing MCQ benchmarks face several criticisms. Older benchmarks like MME~\cite{fu2024mme} and POPE~\cite{data:POPE} are often saturated, while newer, more challenging ones (\emph{e.g.}, MME-RealWorld~\cite{mme_realworld}, RealWorldQA~\cite{data:Realworldqa}) can suffer from ambiguity. Other limitations include answer leakage (mitigated post-MMStar~\cite{zhang2024mmstar}) and data redundancy~\cite{bench:redundancy}. Additionally, there is a recognized deficiency in benchmarks for specific industrial applications and non-English languages, with limited options like MME-RealWorld-CN~\cite{mme_realworld} and MMBench-CN~\cite{liu2023mmbench}.

\section{Dateset Construction}

The images in \textbf{VER-Bench} were sourced from the internet, primarily depicting real-world scenes. Over \textbf{3000} images are collected, with an average resolution of around \textbf{1500 $\times$ 2000 pixels}.

\subsection{Annotation Pipeline and Curation Process}

\begin{figure*}[h]
  \centering
  \includegraphics[width=\linewidth]{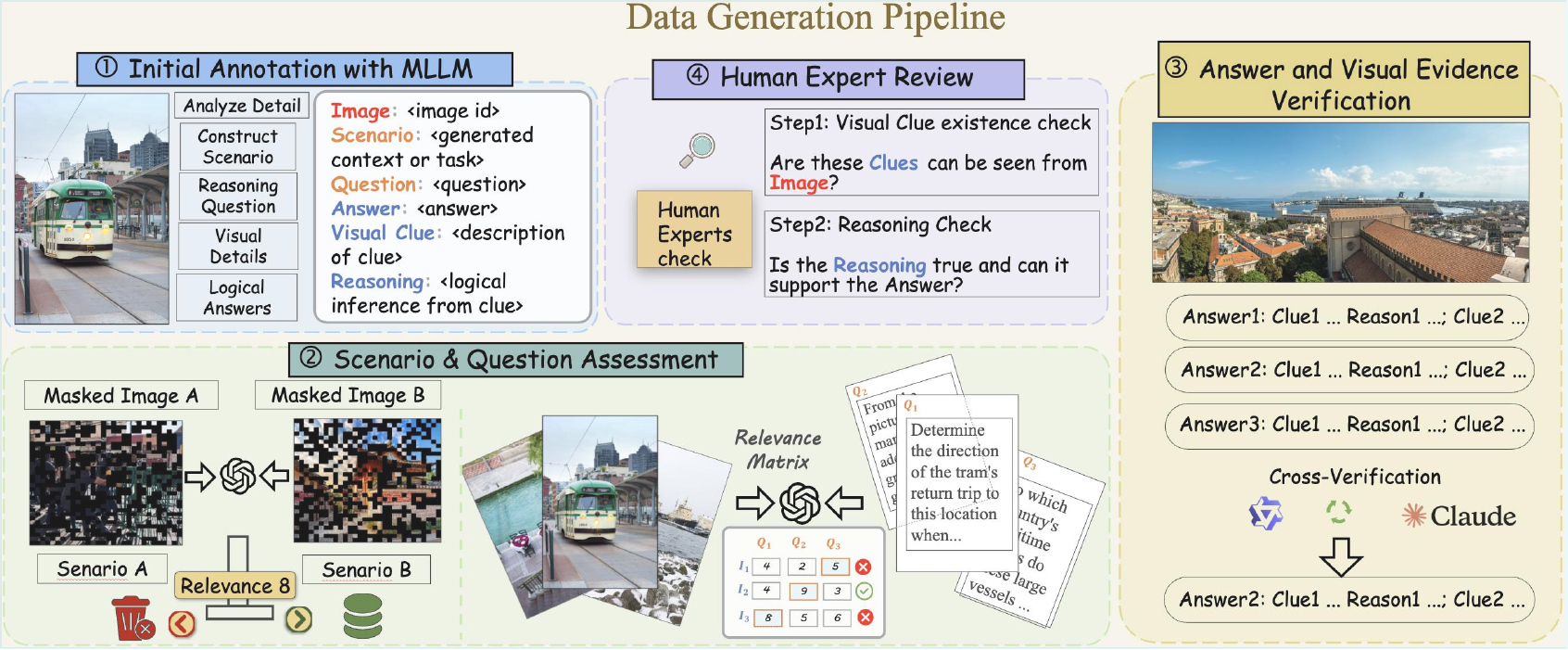}
  \caption{\textbf{The annotation pipeline and curation process.{\mdseries There are four steps, the first step we use a strong MLLM to annotate the relevant information, and then we design two methods to filtering the low quality instances from scenario, question, answer and evidence angle. Last, several human experts check the clue and answer to further guarantee the quality.}}}
  \label{fig:2} 
\end{figure*}

We construct the dataset through a four-stage annotation pipeline (as shown in Fig.~\ref{fig:2}) to generate high-quality visual evidence reasoning chains for real-world scenarios.

\textbf{{Step 1: Initial Annotation.}}
With image as visual input, Gemini-2.5-Pro model is then tasked to: 1. Analyze visual details from the image. 2. Construct a realistic scenario based on the observation. 3. Design a context-driven reasoning question. 4. Provide logical answers supported by visual details.
% The output from this step is structured as shown in Fig.~\ref{fig:2}.

\textbf{{Step 2: Scenario and Question Assessment.}}
To ensure alignment between scenarios and visual content, and the relevance of questions, we employ two validation procedures: 1. GPT-4o evaluates the relevance between the masked image (randomly masked by 50\% area) and the scenario; 2. Multiple random subsets are analyzed using GPT-4o relevance metrics. Entries with lower original image-question relevance compared to alternative pairings are removed.
These strategies ensure that only image-scenario-question pairs with strong mutual relevance are retained.

\textbf{{Step 3: Answer and Visual Evidence Verification.}}
This step ensures the dataset's internal consistency and factual accuracy. We cross-verified answers, reasoning, and supporting visual evidence using multiple MLLMs (\emph{e.g.}, Qwen2.5-VL-72B, Claude-3.5-Sonnet). Only instances with consensus on the final answer and reasoning chain across all models were retained, to minimize model-specific biases and enhancing annotation reliability.

\textbf{{Step 4: Human Expert Review.}}
In two review rounds, human experts ensure accuracy and remove errors: 1. Visual Clue Verification: Experts confirmed the presence of each annotated visual clue in the image, eliminating any incorrect or hallucinated clues; 2. Reasoning Validation: Each reasoning step was reviewed for logical validity and alignment with real-world knowledge.

Additionally, bounding boxes were manually annotated for all visual clues to precisely locate them within the image, enhancing the reasoning chain's interpretability and traceability.

\subsection{Dataset Overview}
\label{subsec:dataset_overview}
The dataset comprises 342 images and 374 Q\&A pairs, with 614 visual clues annotated with 977 bounding boxes. These clues cover an average of 0.25\% of the image area, making them ideal for fine-grained reasoning evaluation. 
VER-Bench is categorized into six tasks (Fig.~\ref{fig:teaser}), detailed below:
\begin{figure*}
  \includegraphics[width=\textwidth]{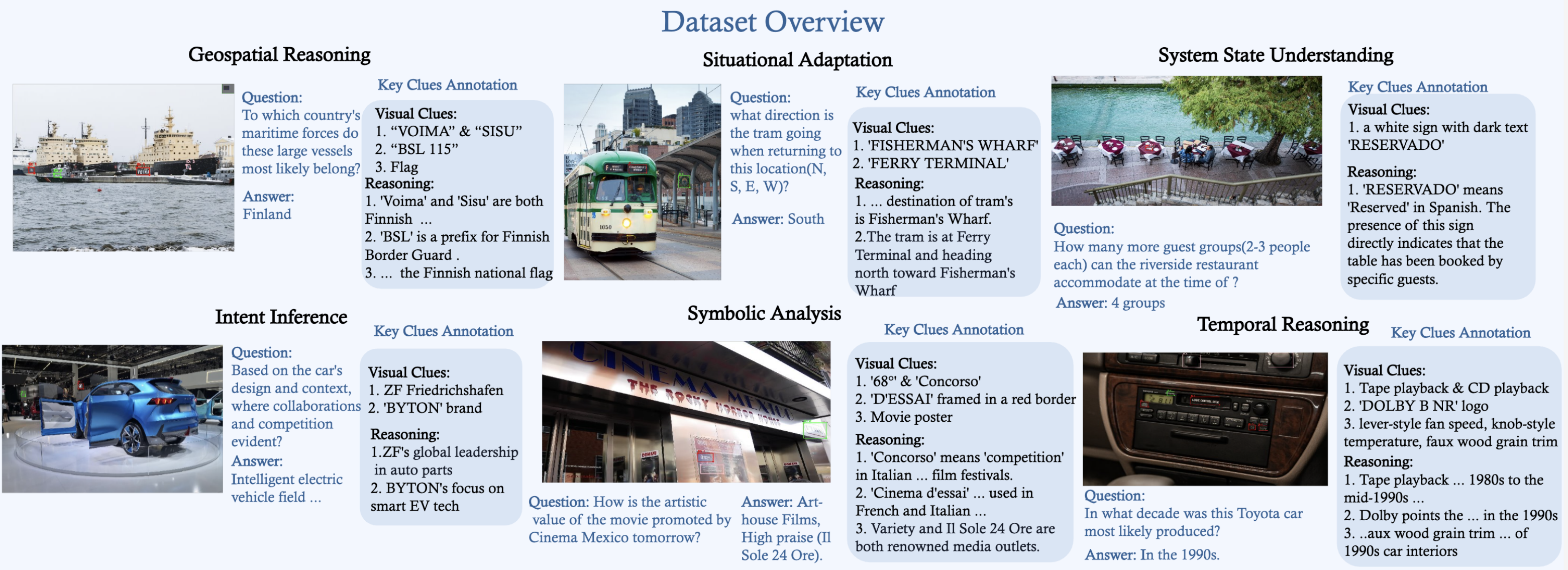}
  \caption{Examples in VER-Bench. {\mdseries Each clue is marked with a box of a different color in the image.}}
  \label{fig:teaser}
\end{figure*}

\begin{figure}[h]
  \centering
  \includegraphics[width=0.9\linewidth]{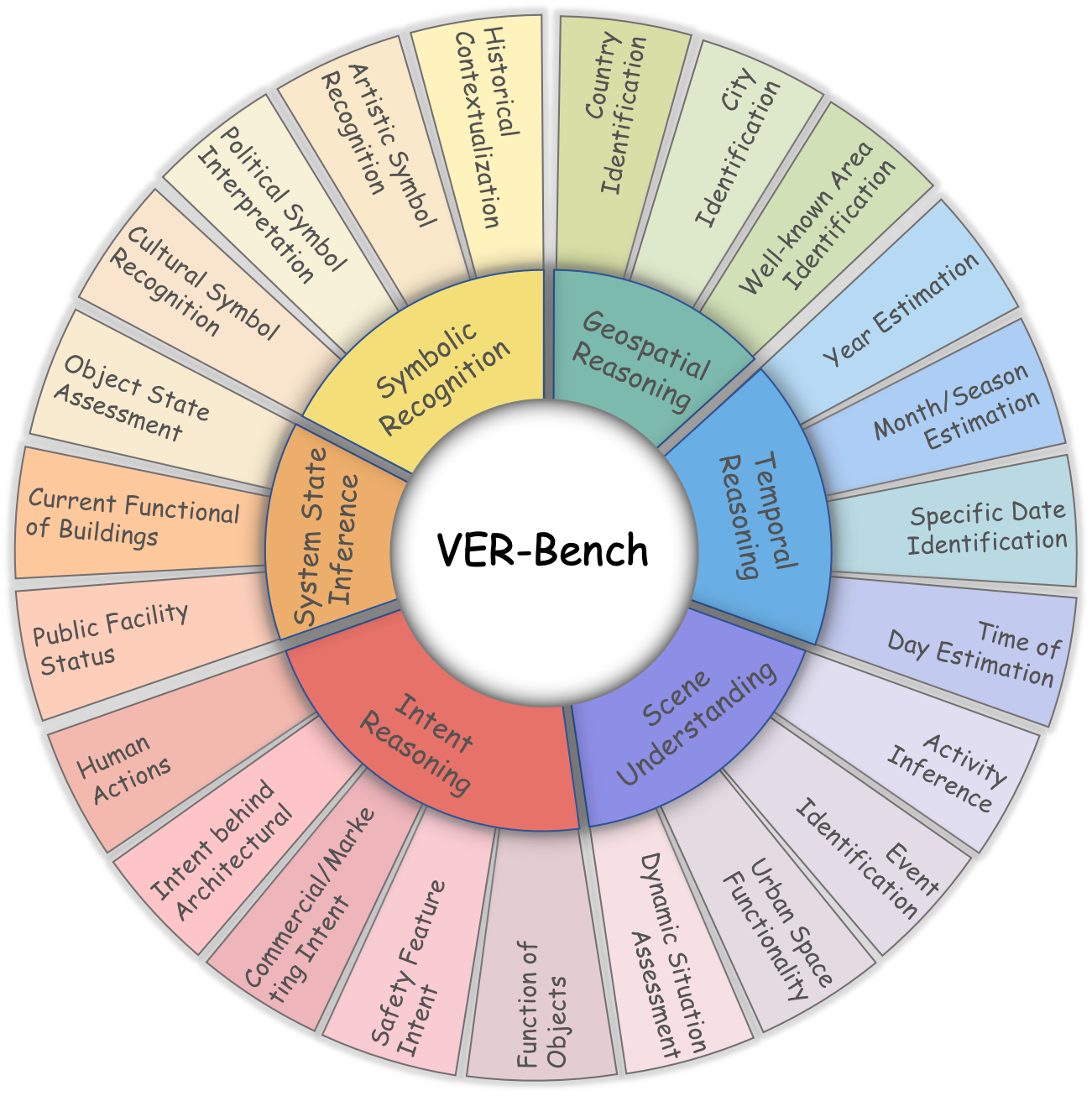}
  \caption{VER-Bench includes 6 tasks covering 23 sub-dimension. {\mdseries A total of 614 visual clues with about 1k boxes are annotated. The average size of image is 1500$\times$2000 and the average size of boxes related to clues is 0.25\% of the whole image.}}
  \label{fig:3} 
\end{figure}

\textbf{Geospatial Reasoning.}  
This task identifies geographic locations using visual clues like landmarks, scripts, symbols, and vehicle details. It requires global geography knowledge and cross-referencing multiple clues for accurate localization.

\textbf{Temporal Reasoning.}  
This task infers time information (\emph{e.g.}, dates) from clues like clocks, text, seasonal signs, and event-specific elements.

\textbf{Situational Adaptation Reasoning.}  
This task interprets complex scenes to infer situational contexts (\emph{e.g.}, events, activities) by understanding human behavior, spatial use, and environmental cues.
    
\textbf{Intent Inference.}  
This task uncovers the underlying purpose of objects, designs, or behaviors, requiring semantic abstraction and contextual interpretation.
    
\textbf{System State Understanding.}  
This task assesses the operational status of facilities or objects by analyzing visual signs like warning labels, crowd flow, or device status.

\textbf{Symbolic Analysis.}  
This task recognizes and interprets cultural or historical symbols (\emph{e.g.}, emblems, icons) using both visual perception and deep cultural knowledge.

\subsection{Evaluation Metric of VER-Bench}
\label{subsec:evaluation_metric}

VER-Bench evaluates models using four metrics: Answer Correctness, Clue Coverage, Reasoning Consistency, and Evidence-Answer Relevance. During evaluation, models are prompted to follow a structured output format (details in Appendix):
$$
\left\{A, \left[(c_1, r_1), \dots, (c_N, r_N)\right]\right\}
$$
where $ A $ represents the final answer, $ c_i $ denotes the $i$-th visual clue identified by the model, and $ r_i $ is the reasoning based on the clue.

The corresponding ground truth is defined as:
\begin{equation}
\left\{A_{gt}, \left[(c_1^{(gt)}, r_1^{(gt)}), \dots, (c_M^{(gt)}, r_M^{(gt)})\right]\right\},
\end{equation}where $ A_{gt} $ denotes the correct answer, and $ c_i^{(gt)} $, $ r_i^{(gt)} $ represent the $i$-th visual clue and its corresponding reasoning from the ground-truth (GT). Each instance is then scored along the following axes:

%using an extra Model (GPT-4o)
    
 \textbf{Answer Correctness (AC):} 
    The AC is determined by the semantic similarity between the model's answer $ A $ and the GT answer $A_{gt}$. The higher $AC$, the greater alignment in meaning and conclusion.
\begin{equation}
\label{eq:ac} 
AC =  \operatorname{LLM}_{\text{similarity}}(A_{gt}, A).
%AC =  \operatorname{LLM}(A_{gt}, A).
\end{equation}

 \textbf{Clue Coverage (CC):}
    This metric evaluates how completely the model captures the information in the ground truth. The higher $CC$ indicates that more critical visual clues are correctly recognized and described by the model.
\begin{equation}
\label{eq:cc} 
CC =  \operatorname{\text{Matching degree}}(C_{gt}, C),
\end{equation}where $C_{gt} = \left\{c_1^{(gt)}, \dots, c_M^{(gt)}\right\}$, $C = \left\{c_1, \dots, c_N\right\}$.
    
 \textbf{Reasoning Quality (RQ):}
    This metric measures how well the model's reasoning consistent with the ground-truth reasoning. A high $RQ$ implies that the model's logic is semantically aligned with the expected reasoning process.
\begin{equation}
\label{eq:rc} 
RC =  \operatorname{\text{Consistency}}(R_{gt}, R),
\end{equation}where $R_{gt} = \left\{r_1^{(gt)}, \dots, r_M^{(gt)}\right\}$, $R = \left\{r_1, \dots, r_N\right\}$.

\textbf{Evidence-Answer Relevance (ER)}: The $ER$ measure whether each reasoning based on visual clue leads to the final answer and whether any reasoning is off-topic or irrelevant.
\begin{equation}
\label{eq:er} 
\quad ER = \frac{\sum_{i=1}^{N} \text{Support Rate}[(c_i, r_i), A]}{N} \times 10.
\end{equation}

The LLM used to score answer correctness, clue coverage, reasoning quality, and evidence-answer relevance is GPT-4. All scores range from 0 to 10. The final average accuracy is computed as:

\begin{equation}
\label{eq:Acc} 
Average\ Accuracy = \frac{1}{10}\sum_{j=1}^{K} (\frac{AC_{j} + CC_{j} + RC_{j}+ ER_{j}}{4}) \times 100\%,
\end{equation}where $K$ is the total number of instances in VER-Bench.

\subsection{Holistic Evidence Evaluation \emph{vs.} Individual Clue-Reasoning Matching}
In $CC$ and $RQ$ metrics, we do not first use LLM to judge whether each model output evidence (clue-reasoning pair) matches the GT evidence and then calculate recall based on matches. Instead, we concatenate all model output clues and GT clues separately and directly evaluate their overall matching score using LLM.

Visual reasoning differs from problems requiring strict sequential logic (\emph{e.g.}, math, programming). In visual reasoning, clue-reasoning pairs often lack strong sequential dependencies, and the order of identification typically does not affect the outcome. These evidence pairs are complementary: identifying more pairs improves answer reliability. Additionally, imperfect alignment between model-output and GT clues results in non-binary matching, complicating recall calculations.

\section{Experiments and Analysis}

In this section, we systematically evaluate state-of-the-art (SOTA) models on VER-Bench. We detail the experiment setup in Sec.~\ref{setup} and report quantitative results and analysis insights in Sec.~\ref{results}.

\subsection{Experiment Setups}\label{setup}
We evaluate 12 open-source models, including Qwen2.5-VL (8B, 32B, 72B)~\cite{bai2023qwenvl}, Qwen2-VL (8B)~\cite{wang2024qwen2}, InternVL3 (8B, 38B, 78B)~\cite{zhu2025internvl3}, InternVL2.5 (8B)~\cite{chen2024expanding}, InternVL2 (8B)~\cite{dong2024internvl2}, MiniCPM-V-2.6, MiniCPM-o-2.6~\cite{yao2024minicpm}, and LLaVA-OneVision-72B~\cite{llava-ov}. Additionally, we evaluate three closed-source models: Gemini-2.5-Pro-Preview~\cite{team2023gemini}, GPT-4o~\cite{achiam2023gpt4}, and Claude-3.7-Sonnet~\cite{anthropic273639283model}.
Open-source models are tested on NVIDIA H20 GPUs using the MS-Swift framework, while closed-source models are evaluated via API calls. GPT-4o is used to score predictions against ground truth.

\setlength{\tabcolsep}{5.5pt}
\begin{table*}
\renewcommand\arraystretch{1}
  \caption{The result of evaluated MLLMs in four metrics, and sorted by Average Accuracy.}
  \label{tab:commands}
  \resizebox{1\textwidth}{!}{
  \begin{tabular}{llccccc}
    \toprule
    \makecell[c]{Model} & \makecell[c]{LLM} & \makecell[c]{Answer \\  Correctness} & \makecell[c]{Clue \\ Coverage} & \makecell[c]{Reasoning \\  Quality} & \makecell[c]{Evidence to \\ Answer Relevance} & {Average Accuracy}\\
    \midrule
    \multicolumn{7}{c}{ \textbf{\textit{Open-Source MLLMs}}} \\ % <--- 这是新加的行
    \specialrule{0.1pt}{\aboverulesep}{\belowrulesep}
    MiniCPM-o 2.6 & Qwen2.5-7B~\cite{yang2025qwen2}& 42.4 & 20.5& 28.8 & 76.0 & 41.9\\
    MiniCPM-V 2.6 & Qwen2-7B~\cite{yang2025qwen3}& 41.0 & 20.2& 27.3& 82.1 & 42.7\\
    LLaVA-OneVision-72B & Qwen2-72B~\cite{yang2025qwen3}& 48.8 & 24.7& 33.7& 86.5 & 48.4\\
    InternVL2-8B & InternLM-7B~\cite{cai2024internlm2}& 35.2 & 16.5& 22.2& 83.9 & 39.4\\
    InternVL2.5-8B & InternLM-7B~\cite{cai2024internlm2}& 38.9 & 19.8& 25.5& 77.5 & 40.4 \\
    InternVL2.5-38B & Qwen2.5-32B~\cite{yang2025qwen2}& 46.6 & 30.0& 35.3& 87.4 & 49.8\\
    InternVL3-8B & Qwen2.5-7B~\cite{yang2025qwen2}& 47.4 & 31.8& 39.0& 83.0 & 50.3\\
    InternVL3-78B & Qwen2.5-72B~\cite{yang2025qwen2}& 55.1 & 40.6& 45.7& 86.2 & 56.9\\
    Qwen2-VL-7B & Qwen2-7B~\cite{yang2025qwen3} & 52.2 & 26.6& 34.3& 79.8 & 48.2\\
    Qwen2.5-VL-7B  & Qwen2.5-7B~\cite{yang2025qwen2}& 52.9 & 29.7& 34.4& 84.8 & 50.4\\
    Qwen2.5-VL-72B & Qwen2.5-72B~\cite{yang2025qwen2}& 61.8 & 38.2& 46.6& 92.0 & 59.6\\
    Qwen2.5-VL-32B & Qwen2.5-32B~\cite{yang2025qwen2}& 63.6 & 49.6& 45.6& 84.4 & 60.8\\
    \midrule 
    \multicolumn{7}{c}{ \textbf{\textit{Closed-Source MLLMs}}} \\ 
    \midrule 
    Claude-3.7-Sonnet-20250219 & \makecell[c]{-}& 52.7 & 37.2& 43.4& 92.1 & 56.3\\
    GPT-4o  & \makecell[c]{-} & 62.8 & 46.7& 54.5& 93.3 & 64.4\\
    Gemini-2.5-Pro-Preview-05-06 & \makecell[c]{-} & 79.0 & 67.9& 69.4& 90.7 & 76.8 \\
    \bottomrule
  \end{tabular}}
\end{table*}

\subsection{Quantitative Results}\label{results}

% Table~\ref{tab:commands} evaluates open-source and closed-source MLLMs on VER-Bench using four metrics: Answer Correctness (AC), Clue Coverage (CC), Reasoning Quality (RQ), Evidence-Answer Relevance (ER).
Table~\ref{tab:commands} evaluates open-source and closed-source MLLMs on VER Bench, using four metrics: Answer Correctness (AC), Clue Coverage (CC), Reasoning Quality (RQ), and Evidence-Answer Relevance (ER). These metrics collectively assess not only the final answer accuracy but also the model’s ability to identify relevant visual clues, perform logical reasoning, and align evidence with responses.

% Closed-source models generally outperform open-source ones, with Gemini-2.5-Pro-Preview achieving the highest Average Accuracy of 76.8, surpassing GPT-4o by 12.5 points. Among open-source models, Qwen2.5-VL-32B achieved the top score of 60.8, outperforming larger models like Qwen2.5-VL-72B and InternVL3-78B. This is attributed to its reinforcement learning strategy, which enhances fine-grained visual analysis and instruction following. Qwen2.5-VL-72B closely follows at 59.6, while InternVL3-78B scored 56.9. LLaVA-OneVision-72B, despite its larger parameters, performs comparably to smaller Qwen-VL and InternVL models, likely due to its focus on cross-modal migration to multi-images and videos rather than single-image understanding. The edge models MiniCPM-V 2.6 and MiniCPM-o 2.6 show strong performance for their size, surpassing InternVL2-8B and InternVL2.5-8B.

Closed-source models lead in performance, with Gemini-2.5-Pro-Preview scoring 76.8 average accuracy, surpassing GPT-4o by 12.5 points. Among open-source models, Qwen2.5-VL-32B tops at 60.8, outperforming larger variants due to its reinforcement learning strategy that improves visual detail understanding and instruction following. Qwen2.5-VL-72B (59.6) and InternVL3-78B (56.9) trail closely. LLaVA-OneVision-72B performs on par with smaller models, likely because it prioritizes multi-image and video tasks over single-image reasoning. Lightweight models MiniCPM-V 2.6 and MiniCPM-o 2.6 also punch above their weight, outperforming larger 8B-class models.

\subsubsection{Answer Correctness} This metric assesses the semantic similarity between the model-generated final answer and the GT answer. Higher $AC$ scores indicate greater alignment in meaning and conclusion. Gemini-2.5-Pro-Preview achieves the highest $AC$, demonstrating its proficiency in generating semantically accurate answers. Qwen2.5-VL-32B (61.8) and GPT-4o (62.8) also show strong performance. Conversely, models like InternVL2-8B (35.2) have significantly lower $AC$, indicating difficulties in reaching the correct conclusion despite potential strengths in other aspects (\emph{e.g.}, clue identification or reasoning). This metric is crucial as it directly reflects the model's ability to fulfill the primary task objective.

\subsubsection{Clue Coverage} Clue Coverage ($CC$) measures how well models identify GT visual clues. Higher $CC$ indicates better recognition of critical visual information, leading to more accurate localization of relevant image regions and higher answer accuracy. For example, Qwen-VL-series models saw $CC$ increase from 26.6 (Qwen2-VL-7B) to 38.2 (Qwen2.5-VL-72B), correlating with improved answer correctness. However, many lower-performing open-source models (\emph{e.g.}, InternVL2-8B with 16.5 $CC$) struggle significantly, often generating excessive irrelevant clues. This redundancy disrupts the correlation assessment with GT clues, likely due to poor instruction-following. Low $CC$ suggests models overlook essential visual details, impairing reasoning and answer correctness, thus highlighting the importance of effective visual perception in MLLMs.

\subsubsection{Reasoning Quality}

% $RQ$ measures the semantic consistency between the reasoning generated by the model and the GT reasoning. A high $RQ$ value signifies that the model's logical deduction process closely aligns with the expected correct reasoning process. Gemini-2.5-Pro-Preview leads with 69.4 points, demonstrating its ability not only to identify clues correctly but also to explain how these clues lead to conclusions in a manner akin to human standards. GPT-4o (54.5) and Claude-3.7-Sonnet (43.4) also exhibit strong reasoning quality. Among open-source models, Qwen2.5-VL-72B (46.6), InternVL3-78B (45.7), and Qwen2.5-VL-32B (45.6) stand out, indicating they excel at simulating correct reasoning paths compared to other open-source counterparts. In contrast, models such as InternVL2.5-8B (25.5) and InternVL2-8B (22.2) score lower. This suggests their reasoning processes may contain logical flaws, insufficient explanations, or significant semantic deviations from GT reasoning---even if they identify some correct visual clues.

$RQ$ measures how semantically consistent the model’s reasoning is with the ground truth. A higher $RQ$ score indicates closer alignment with correct logical deduction. Gemini-2.5-Pro-Preview leads at 69.4, showing strong human-like reasoning. GPT-4o (54.5) and Claude-3.7-Sonnet (43.4) also perform well. Among open-source models, Qwen2.5-VL-72B (46.6), InternVL3-78B (45.7), and Qwen2.5-VL-32B (45.6) stand out, demonstrating effective reasoning. In contrast, models such as InternVL2.5-8B (25.5) and InternVL2-8B (22.2) score lower. This suggests their reasoning processes may contain logical flaws, insufficient explanations, or significant semantic deviations from GT reasoning---even if they identify some correct visual clues.

\subsubsection{Evidence-to-Answer Relevance}

Across all metrics, the average score for $ER$ is significantly higher than others, with even the lowest score exceeding 76. This indicates that even if a model's final answer ($AC$) or reasoning process ($RQ$) deviates from the GT, it can still construct internally self-consistent "evidence-reasoning-answer". For instance, Qwen2.5-VL-72B outperformed closed-source models like Gemini-2.5-Pro-Preview in $ER$, highlighting its potential for generating coherent explanations. Conversely, lower $ER$ scores, such as MiniCPM-o 2.6 (76.0) and InternVL2.5-8B (77.5), suggest that these models' reasoning steps may be more divergent or contain redundant information with limited support for their conclusions. $ER$ is crucial for evaluating the coherence and persuasiveness of model outputs. A high $ER$ score signifies a more rigorous and focused reasoning process.

$RQ$ and $ER$ together show how well a model can think and if its thinking is consistent. High $RQ$ means the model can copy good thinking. High $ER$ means the model can organize its thoughts to back up its conclusions, even if they aren't perfect. Gemini-2.5-Pro-Preview-05-06 excels in both metrics, demonstrating strong logical reasoning and consistency. In contrast, open-source models generally perform adequately in $ER$ but exhibit a significant gap in $RQ$ compared to top closed-source models. This discrepancy suggests that future research should focus on enhancing the logical coherence of reasoning in open-source models, bringing them closer to human-like reasoning patterns. Overall, these two metrics provide valuable insights into the reasoning capabilities and limitations of MLLMs when tackling complex tasks.

\section{Conclusion}
In this paper, we introduce VER-Bench, a benchmark tailored to evaluate the visual reasoning capabilities of MLLMs. It challenges models to identify critical visual details, interpret their meanings, and integrate fragmented information to derive reliable answers. This process mirrors human visual analysis, where identifying abundant supporting evidence is crucial for strengthening conclusions. Our evaluation results demonstrate a positive correlation between Clue Coverage and Answer Correctness, validating our motivation. VER-Bench represents a significant advancement in the evaluation and development of MLLMs. By promoting nuanced, evidence-driven reasoning, it encourages models to identify and integrate fine-grained visual clues, thereby forming robust conclusions. This framework not only pushes models beyond surface-level understanding but also drives the development of more sophisticated reasoning capabilities. Through VER-Bench, we aim to foster the evolution of MLLMs that approach visual reasoning tasks with the depth and precision characteristic of human cognition.
\begin{acks}
Acknowledgements: This work was supported in part by the Key
Deployment Program of the Chinese Academy of Sciences, China under Grant KGFZD145-25-39 and the Strategic Priority Research Program of Chinese Academy of Sciences under Grant E1XA310103.
\end{acks}

\bibliographystyle{ACM-Reference-Format}
\bibliography{sample-base}

%\appendix


\end{document}